\documentclass[conference]{IEEEtran}
\ifCLASSINFOpdf
\else
\fi

\usepackage{import}
\usepackage{mathtools, amssymb, amsthm,commath}
\usepackage{paralist}
\usepackage{todonotes}

\usepackage{soul,listings}
\usepackage{xcolor}

\definecolor{xcgreen}{rgb}{0.24,0.52,0}
\definecolor{mypurple}{rgb}{0.73,0.17,0.65}
\definecolor{wbi_position}{rgb}{0.51,0.85,0.67}
\definecolor{wbi_velocity}{rgb}{0.51,0.67,1.0}
\definecolor{wbi_torque}{rgb}{0.83,0.58,0.60}

\usepackage{algpseudocode}
\usepackage{algorithm}

\begin{document}
\title{A Whole-Body Software Abstraction layer for Control Design of free-floating Mechanical Systems}

% \author{\IEEEauthorblockN{Francesco Romano, Silvio Traversaro and Francesco Nori}
% \IEEEauthorblockA{iCub Facility Department\\
% Istituto Italiano di Tecnologia, Genova, Italy\\
% Email: name.surname@iit.it}
% }


\author{\IEEEauthorblockN{Francesco Romano\IEEEauthorrefmark{1},
Silvio Traversaro\IEEEauthorrefmark{1},
Daniele Pucci\IEEEauthorrefmark{1},
Jorhabib Eljaik\IEEEauthorrefmark{2},
Andrea Del Prete\IEEEauthorrefmark{3}, and
Francesco Nori\IEEEauthorrefmark{1}}
\IEEEauthorblockA{\IEEEauthorrefmark{1}iCub Facility Department,
Istituto Italiano di Tecnologia, Genova, Italy. Email: name.surname@iit.it}
\IEEEauthorblockA{\IEEEauthorrefmark{2}Institut des Syst\`{e}mes Intelligents et de Robotique, UPMC Univ. Paris, 06, Paris, France. Email: eljaik@isir.upmc.fr}
\IEEEauthorblockA{\IEEEauthorrefmark{3}LAAS/CNRS, Toulouse, France. Email: adelpret@laas.fr}}


\maketitle

\begin{abstract}
% The complexity of modern days redundant robots calls for scalable and flexible softwaresolutions.
% Modern day redundant robots, such as legged robots, are complex platforms that can achieve multiple objectives at the same time.
% % Whole-Body control is a control approach in which, during the execution of a single objective, the whole robot together with all the possible constraints and other tasks are taken into account.
% The use of the Task function approach has become popular in the whole-body control context since it allows to specify both the control objectives and the constraints the robot is subject to.
% As a consequence, software frameworks for task-based whole-body control have gained popularity in the robotics community.
% While these frameworks allow one to easily specify objectives and constraints, they often bind the user to the controller architecture already implemented in the library.
In this paper, we propose a software abstraction layer to simplify the design and synthesis of whole-body controllers without making any preliminary assumptions on the control law to be implemented.
The main advantage of the proposed library is the decoupling of the control software from implementation details, which are related to the robotic platform.
Furthermore, the resulting code is more clean and concise than ad-hoc code, as it focuses only on the implementation of the control law.
In addition, we present a reference implementation of the abstraction layer together with a Simulink interface to provide support to Model-Driven based development.
We also show the implementation of a simple proportional-derivative plus gravity compensation control together with a more complex momentum-based bipedal balance controller.
% the results achieved by using our library on a more complex momentum-based balancing controller.
\end{abstract}


% \fi
\IEEEpeerreviewmaketitle

%!TEX root = ../wbi.tex
\section{Introduction}

Nowadays, robotics is moving from the original industrial context to more human-like environments.
Foreseen applications involve robots with augmented autonomy and physical mobility. 
Within this novel context, physical interaction influences stability and balance.
Consequently, the requirements and tasks that we expect from some platforms are changing as well. 
Instead of precise positioning tasks confined in cages in industrial assemblies, robots are foreseen to help in everyday-life tasks such as cleaning houses or elderly assistance. 

The increase in complexity of robotic systems demands an increase in complexity of the corresponding control software.
While ad-hoc solutions can be easy to implement, it is important to consider scalability, flexibility and portability of the developed software.
The possibility to use the same software to control more than one platform can be of enormous importance in simplifying the testing, tuning, and deployment of the same controller on different robots.

% Foreseen applications involve robots with augmented autonomy and physical mobility.
% Within this novel context, physical interaction influences stability and balance.
Whole-body control has received an increased attention by the robotics community because of the possibility it offers to accomplish tasks coordination and to fully take advantage of the robots dynamics in presence of contacts.
% To allow robots to surpass barriers between interaction and posture control, CoDyCo will be grounded in principles governing whole-body coordination with contact dynamics.
% In the context of multiple degrees of freedom redundant robots, whole-body control has received an increased attention by the robotics community.
Indeed, the possibility to specify multiple objectives, even conflicting, at the same time opens the possibility to properly exploit robots for complex scenarios.
% Unfortunately the existence of multiple objective to be satisfied greatly complexifies the control problem.
% Whole-body control is a control formulation in which the whole robot is taken into account during the execution of a single task, with all the possible constraints and by considering all the tasks the robot should execute.
In particular, citing the definition from the RAS Technical Committee \cite{RAS_WBC_TC} ``\emph{Whole-Body Control aims to i) define a small set of simple, low-dimensional rules (e.g., equilibrium, self collision avoidance, etc.) ii) that are sufficient to guarantee the correct execution of any single task, whenever feasible [...], and of simultaneous multiple tasks [...], iii) exploiting the full capabilities of the entire body of redundant, floating-based robots in compliant multi-contact interaction with the environment}''.

In the context of whole-body control the Task function approach \cite{Samson1991} has been successfully used. 
In this method, the control objectives are represented as $n$-dimensional continuous output functions, called tasks, to be regulated to zero.
All the tasks, together with possible constraints are then transformed into a constrained optimization problem. % to be solved by an appropriate solver.
From a software perspective, different implementations exist nowadays, among which the Stack of Task (SoT)~\cite{Mansard2009}, OpenSoT~\cite{rocchi2015opensot}, ControlIt!~ \cite{fok2016controlit} and the Instantaneous Task Specification using Constraints (iTaSC)~\cite{DeSchutter2007}.
The above softwares allow the user to specify the objectives and constraints but they solve the control problem internally. 
A disadvantage is that they force the user to choose a specific task-based approach to obtain the control solution thus denying the control designer the possibility to synthesize different control laws.

In this paper we propose a different approach for the whole-body control of mechanical systems.
We deal with the control problem from a more general perspective, without limiting the user to the use of a task-based approach.
Indeed, when we consider a generic control system, we usually identify three main building blocks:
% A control system has to access, in general, three basic types of information of the controlled plant:
\begin{itemize}
    \item Plant model. If we consider a model-based control system, in this block the information about the plant model given the current plant state are computed.
    \item Feedback from the plant. This usually implies the possibility to obtain the current state of the controlled plant.
    \item Actuation. The control system must interact with the plant.
\end{itemize}
Any control-oriented software library must provide the above features to be of any use.
Given the complexity of robotic systems it can be difficult, time consuming and error prone to write the controller directly in a low-level programming language such as C++. 
Nevertheless the control library must be efficient as it is usually required to have fast control loops. 
% I don't think citing 100 Hz is useful here. 
% For example a commonly chosen control loop frequency is $100 \mathrm{Hz}$, but this can greatly vary depending on the dynamic response of the considered controlled system.
% Support to model-driven engineering is thus an additional welcome functionality of the library.
The aforementioned requirements serve as motivation for a model-based driven approach in such control libraries.

In this paper we propose a software abstraction layer which is responsible of decoupling the control software from 
\begin{inparaenum}[i)]
    \item the actual interface used to obtain the state feedback;
    \item the actual interface used to command the actuation;
    \item the dynamic software library used to represent the robot dynamical model.
\end{inparaenum}
Furthermore the proposed library is scalable and easily portable to other robots or different configurations.

% - Flexibility: test in simulation, on the robot, etc

This paper is structured as follows. Section~\ref{sec:background} introduces the mathematical formulation of the dynamics of mechanical systems and it shows an example of a simple classic controller.
Section~\ref{sec:software_architecture} describes the architecture of the proposed whole-body abstraction library and its key elements.
A specific implementation is instead presented in Section~\ref{sec:software_implementation}.
The controller mathematically introduced in Section~\ref{sec:background} is implemented with the proposed library in Section~\ref{sec:experiments}. 
Finally Section~\ref{sec:conclusions} draws the conclusions.

%!TEX root = ../wbi.tex
\section{Dynamics of a Mechanical System}
\label{sec:background}

This section introduces the mathematical formulation commonly used in the robotics literature  to describe the dynamics of mechanical systems, such as robots.
Because a precise formulation of the mathematical problem is out of the scope of the present paper, we refer the interested reader to books on dynamics of mechanical systems \cite{Siciliano2009,Featherstone2007,Murray1994} and control systems \cite{Isidori1995,khalil2002} for further readings.

\subsection{Notation}
Throughout the section we will use the following definitions:
\begin{itemize}
    % \item $\mathbb{R}$ denotes the set of real numbers and $e_i \in \mathbb{R}^m$ is the canonical vector, consisting of all zeros but the $i$-th component which is one.
    \item $\mathcal{I}$ denotes an inertial frame, with its $z$ axis pointing against the gravity. %We denote with $g$ the gravitational constant.
    \item $1_n \in \mathbb{R}^{n \times n}$ is the identity matrix of size $n$; $0_{m \times n} \in \mathbb{R}^{m \times n}$ is the zero matrix of size $m \times n$ and $0_{n } = 0_{n \times 1}$.
    \item Given two orientation frames $A$ and $B$, and vectors of coordinates expressed in these orientation frames, i.e. $\prescript{A}{}p$ and $\prescript{B}{}p$, respectively, the rotation matrix 
    $\prescript{A}{}R_B$ is such that $\prescript{A}{}p = \prescript{A}{}R_B  \prescript{B}{}p$. 
    \item We denote with $S(x) \in \mathbb{R}^{3 \times 3}$ the skew-symmetric matrix such that $S(x)y = x \times y$, where $\times$ denotes the cross product operator in $\mathbb{R}^3$. 
\end{itemize}

\subsection{System modelling}
\label{sec:model}
We assume that the mechanical model is composed of $n+1$ rigid bodies -- called links -- connected by $n$ joints with one degree of freedom each. In addition, we also assume   that the multi-body system is \emph{free floating}, i.e. none of the links has an \emph{a priori} constant pose with respect to the inertial frame. This implies that  the multi-body system possesses $n~+~6$ degrees of freedom. The 
configuration space of the multi-body system can then be characterized by the \emph{position} and the \emph{orientation} of a frame attached to a robot's link -- called 
\emph{base frame} $\mathcal{B}$ -- and the joint configurations. 
More precisely, the robot configuration can be represented by the 
triplet 
\[q = (\prescript{\mathcal{I}}{}p_{\mathcal{B}},\prescript{\mathcal{I}}{}R_{\mathcal{B}},q_j),\] where $(\prescript{\mathcal{I}}{}p_{\mathcal{B}},\prescript{\mathcal{I}}{}R_{\mathcal{B}})$ denotes the origin  and orientation of the \emph{base frame} expressed in the inertial frame, and $q_j$ denotes the \emph{joint angles}. 

% More precisely, the robot configuration space  is defined by
% \begin{equation*}
%     \mathbb{Q} = \mathbb{R}^3 \times SO(3) \times \mathbb{R}^n.
% \end{equation*}
% An element of the set $\mathbb{Q}$ is then a
% It is possible to define an operation associated with the set $\mathbb{Q}$ such that this set is a group. More precisely, given two elements $q$ and $\rho$ of the configuration space, the set $\mathbb{Q}$ is a group under the following operation:
% \begin{IEEEeqnarray}{RCL}
% \label{eqn:groupOperation}
% q \cdot \rho = (p_q + p_\rho, R_q R_\rho, q_j + {\rho}_j).
% \end{IEEEeqnarray}
% Being the direct product of Lie groups, the set $\mathbb{Q}$ is itself a Lie group.
The \emph{velocity} of the multi-body system can then be characterized 
% by the \emph{algebra} $\mathbb{V}$ of $\mathbb{Q}$ defined by:
%     $\mathbb{V} = \mathbb{R}^3 \times \mathbb{R}^3 \times \mathbb{R}^n$.
% An element of $\mathbb{V}$ is then a
by the triplet 
\[\nu = ( ^\mathcal{I}\dot{ p}_{\mathcal{B}},^\mathcal{I}\omega_{\mathcal{B}},\dot{q}_j),\]
 where $^\mathcal{I}\omega_{\mathcal{B}}$ is the angular velocity of the base frame expressed w.r.t. the inertial frame, i.e. $^\mathcal{I}\dot{R}_{\mathcal{B}} = S(^\mathcal{I}\omega_{\mathcal{B}})^\mathcal{I}{R}_{\mathcal{B}}$. 
% It is worth noting that a common choice in the robotics literature is to choose $\mathbb{Q} = SE(3) \times \mathbb{R}^n$, but this would have resulted in a different choice for the base velocity, i.e.
% the first element of $\nu$ would not have been ${}^\mathcal{I}\dot{p}_B$.

% Although the above digression on the robot configuration space may sound pedantic and marginal, let us observe that the choice of the group operation in~\eqref{eqn:groupOperation} implies that an element $\nu \in \mathbb{V}$ is composed of  $\dot{p}$, i.e. the time derivative of the origin of the floating base frame. Other choices for the group operation would imply a different algebra and, consequently, a different representation of the system's \emph{velocity}.

We also assume that the robot is interacting with the environment through $n_c$ distinct contacts. 
The application of the Euler-Poincar\'e formalism \cite[Ch. 13.5]{Marsden2010}
% \footnote{The Euler-Lagrange's formulation can be applied only to mechanical systems evolving in vector spaces. The Euler-Poincar\'e equations, instead, are valid for mechanical systems evolving in arbitrary Lie groups.}
to the multi-body system  yields the following equations of motion: 
\begin{align}
    \label{eq:system_dynamics}
       {M}(q)\dot{{\nu}} + {C}(q, {\nu}) {\nu} + {G}(q) =  B \tau + \sum_{k = 1}^{n_c} {J}^\top_{\mathcal{C}_k} f_k
\end{align}
where ${M} \in \mathbb{R}^{n+6 \times n+6}$ is the mass matrix, ${C} \in \mathbb{R}^{n+6 \times n+6}$ is the Coriolis matrix and ${G} \in \mathbb{R}^{n+6}$ is the gravity term.
$\tau$ are the internal actuation torques and $B$ is a selector matrix which depends on the available actuation, e.g. in case all joints are actuated it is equal to $B = (0_{n\times 6} , 1_n)^\top$.
$f_k = [F_i^\top, \mu_i^\top]^\top \in \mathbb{R}^6$, with $F_i, \mu_i \in \mathbb{R}^3$ respectively the force and corresponding moment of the force, denotes an external wrench applied by the environment on the link of the $k$-th contact.
 % We assume that the application point of the external wrench is associated with a frame $\mathcal{C}_k$, which is attached to the robot's link where the wrench acts on, and has its $z$ axis pointing as the normal of the contact plane. Then,  the external wrench $f_k$ is expressed in a frame whose orientation coincides with that of the inertial frame $\mathcal{I}$, but whose origin is the  origin of $\mathcal{C}_k$, i.e. the application point of the external wrench $f_k$.
The Jacobian ${J}_{\mathcal{C}_k}= {J}_{\mathcal{C}_k}(q)$ is the map between the robot velocity ${\nu}$ and the linear and angular velocity \[ ^\mathcal{I}v_{\mathcal{C}_k} := (^\mathcal{I}\dot{ p}_{\mathcal{C}_k},^\mathcal{I}\omega_{\mathcal{C}_k})\] of the frame $\mathcal{C}_k$, i.e.
\begin{align*} 
^\mathcal{I}v_{\mathcal{C}_k} = {J}_{\mathcal{C}_k}(q) {\nu}.
\end{align*}
% The Jacobian has the following structure.
% \begin{IEEEeqnarray}{RCLRLL}
% \label{eqn:jacobian}
% {J}_{\mathcal{C}_k}(q) &=& \begin{bmatrix} {J}_{\mathcal{C}_k}^b(q) & {J}_{\mathcal{C}_k}^j(q)\end{bmatrix} &\in& \mathbb{R}^{6\times n+6}, \IEEEyessubnumber \\
%  {J}_{\mathcal{C}_k}^b(q) &=&
%  \begin{bmatrix}
%  1_3 & -S(\prescript{\mathcal{I}}{}p_{\mathcal{C}_k}-\prescript{\mathcal{I}}{}p_{\mathcal{B}})\\
%  0_{3\times3} & 1_3 \\
%  \end{bmatrix} &\in& \mathbb{R}^{6\times6} . \IEEEyessubnumber
% \end{IEEEeqnarray}

% Lastly, we assume that  rigid contacts may occur between the robot and the environment.
% The constraint associated with the rigid contact is  modelled as a kinematic constraint that forbids any motion of the frame $\mathcal{C}_k$, i.e. ${J}_{\mathcal{C}_k}(q) {\nu} = 0$.

\subsection{Control Example} % (fold)
\label{sub:control_examples}

To illustrate the use of the dynamical model presented in Section \ref{sec:model} we present the classic Proportional Derivative (PD) plus Gravity compensation controller as example.
% \todo[inline]{ST: I would cut the balancing part altogether, the mathematical details are not useful to understand it when it is referenced in experiments and it hides the point of the paper.}
% The second one, instead, leverages the full dynamical model of the robot and represent the torque-controlled balancing controller currently implemented on the iCub humanoid robot.

% \subsubsection{PD plus Gravity compensation} % (fold)
% \label{ssub:pd_plus_gravity_compensation}
%
This kind of controller has been usually applied to fully-actuated fixed-base robots.
Considering the model presented in Section \ref{sec:model} this means that the base frame position and orientation are constant and known a-priori and thus they are not part of the robot state.
% \todo[inline]{ST: This equation (and the one afterwards) are contradicting the equations in section 2.B . We can either define (q,$\nu$) as the generalized state of the system in section 2.B (without specifying that is composed by a base/joint part) or just zap this redefinition of (q,$\nu$). I vote for the latter.}
% Because of the fixed-base assumption, the robot configuration vector
% \[q \equiv q_j \in \mathbb{R}^n. \]
% Furthermore the velocity and acceleration of the system directly corresponds to respectively the first and second time derivative of the joint configuration, i.e.
% \begin{align*}
%     \nu & \equiv \dot{q}_j \equiv \dot{q} \\
%     \dot{\nu} & \equiv \ddot{q}_j \equiv \ddot{q}.
% \end{align*}
% As a consequence, because of the fully-actuated hypothesis, the selector matrix \[B \equiv 1_n.\]

The control objective is the asymptotical stabilization of a desired constant joint configuration $q_j^d$ or equivalently the asymptotical stabilization to zero of the error
\[\tilde{q}_j := q_j - q_j^d. \]
The choice of the following control action
\begin{equation}\label{eq:pd_plus_grav_law}
    \tau = G_j(q) - K_p \tilde{q}_j - K_d \dot{q}_j
\end{equation}
where $K_p, K_d \in \mathbb{R}^{n\times n}$ are the positive definite proportional and derivative gain matrices and $G_j(q) = [0_{n \times 6}
~1_n]~G(q)$, 
satisfy the control objective, i.e. the stabilization to zero of $\tilde{q}_j$, and it can be proved by Lyapunov arguments \cite[Sec. 6.5.1]{Siciliano2009}.

\section{Software Architecture} % (fold)
\label{sec:software_architecture}

\begin{figure*}[t]
  \centering
    \includegraphics[width=\textwidth]{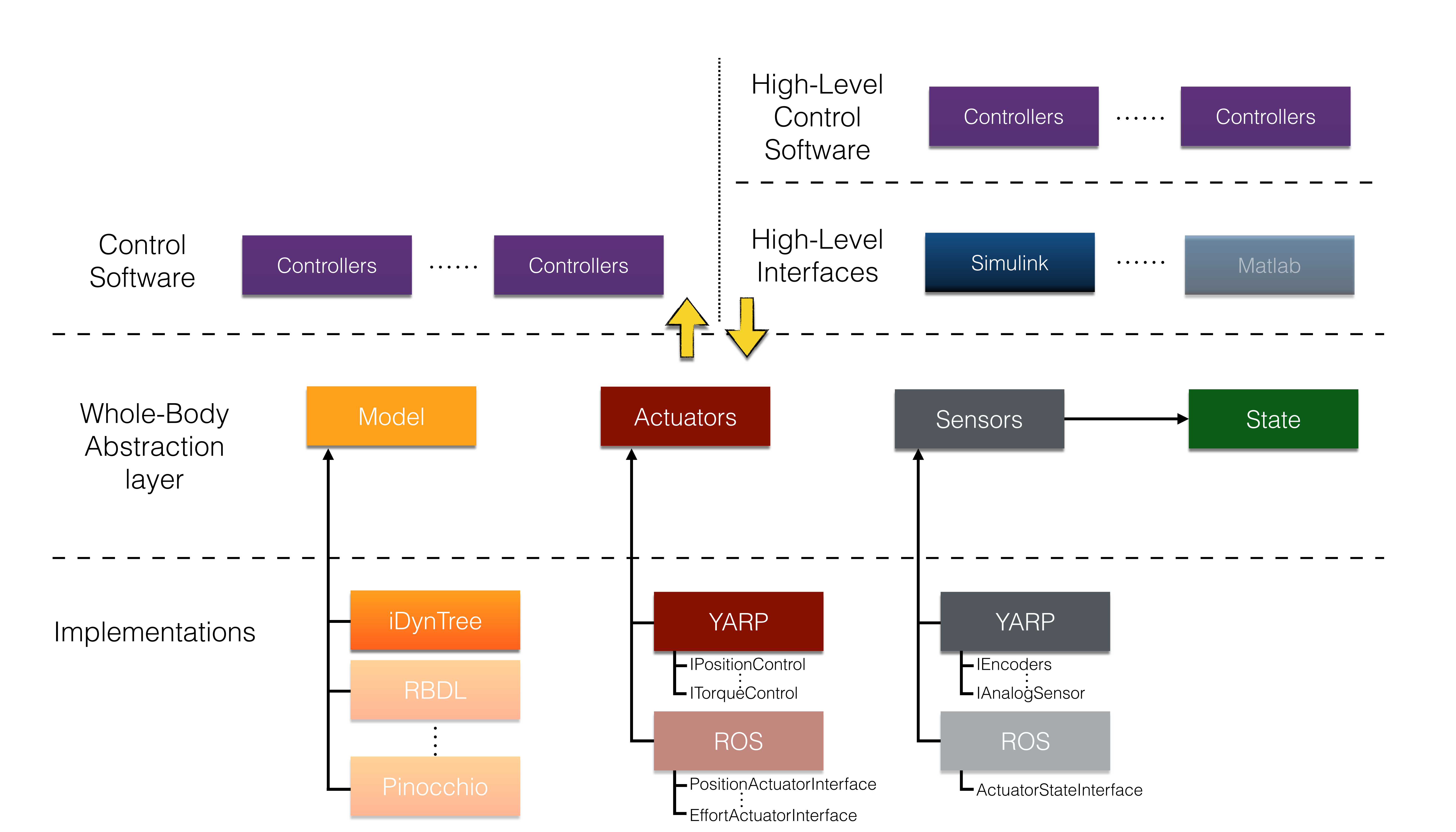}
  \caption{Schema of the proposed whole-body Abstraction layer. Controllers communicate to the kinematic and dynamic library and to the robot software through the proposed library. The implemented elements are drawn with full colour while the shaded blocks represented examples of possible implementations currently unavailable}
  \label{fig:sw_architecture}
\end{figure*}

% \todo[inline]{``Component'' is a quite overloaded concept in Robotics Software Engineering. See
% \cite{brugaliComponent2009,brugaliComponent2010} . Typically it indicates the minimum "block" of computations available in a given robotics framework, for example OROCOS components or ROS Nodes. YARP does not have a real "component", in several aspects both a YARP module or a YARP device could be considered a "Component". Anyhow I don't think we can call this different parts of the libraries "Component".  }

We propose a software abstraction layer to simplify creating whole-body controllers for highly redundant mechanical systems.
Given the requirements introduced in the previous sections we highlight four main elements that must be present in the library, i.e. \emph{Actuators}, \emph{Sensors},  \emph{State} and \emph{Model}.
The abstraction offered by the library allows one to also easily implement higher-level interfaces such as a Simulink\textsuperscript{\textregistered} interface.
Figure~\ref{fig:sw_architecture} summarizes the whole software architecture.

Note that the paradigm described by this library does not assume any particular robot operating system or underlining software as this is left to the actual library implementation.

% \subsection{Library Core} % (fold)
% \label{sub:library_core}
%
% \todo[inline]{I do not understand if the paper until know refers to a generic "decomposition" of the functionality necessary for whole body control (that could be applied to different library/frameworks) or specifically to our WBI stuff. The first hypothesis is nice (if we are able to write it down nicely) but in that case I don't understand why we write that the library should be written in C++. A good example that does that (discussing "general" things and then provide a concrete example) even if in a more complex way is \cite{vanthienen20145c}. }
%
% The core of the library is provided by an abstraction layer implemented in C++.
A crucial feature of the proposed abstraction layer is related to the ordering of the information provided from and to the robot.
In fact, the elements that directly interface the hardware, i.e. the \emph{Actuators}, \emph{Sensors} and \emph{State} have to represent the information in a robot-dependent suitable way.
On the other hand, the \emph{Model} element usually interfaces with libraries that represent the information with the formalism of Eq.~\eqref{eq:system_dynamics}.
To further complicate the problem, the control software may want to access only a subset of the degrees of freedom modelled by the dynamics library, or provided by the robot.
The whole-body abstraction library must thus orchestrate all the various elements to provide a unified interface to the control software.

We now describe in detail the role that each element has in the proposed library.

\subsection{Actuators} % (fold)
\label{sub:actuators}

The \emph{actuators} element abstracts the actual control of the robot motors.
In particular it exposes the possible motors controllable mode, e.g. position control, velocity control and torque control just to cite the most common.
Of course, it also provides the possibility to specify the references for the low level controllers.

% subsubsection actuators (end)

\subsection{Sensors} % (fold)
\label{sub:sensors}

The \emph{sensors} element is the counterpart of the \emph{actuators} element.
In fact, it abstracts all the sensors available on the robot, usually the readings from encoders, force/torque sensors or accelerometers and it is responsible for providing access to the latest sensor measurements.

% subsubsection state (end)

\subsection{State} % (fold)
\label{sub:state}

The \emph{state} element represents all the possible information which can be measured or estimated on the robot.
This implies that \emph{state} encompasses the information provided by the \emph{sensor} element.
Furthermore, it provides additional information which can come from estimation or filtering of the data.
For example, if the robot provides only joint position measurements, e.g. coming from the joint encoders, a first and second derivative filter can provide velocity and acceleration measurements. 
In case this information is provided by the robot itself, no additional processing is required from the interface.
It is important to notice that in both cases the control software using the abstraction library will remain exactly the same.

% subsubsection state (end)

\subsection{Model} % (fold)
\label{sub:model}

The last element is the \emph{model} element.
It abstracts the kinematic- and dynamic-related information that a controller needs while computing the control law.
In general data are represented with the formalism of Eq.~\eqref{eq:system_dynamics}.
Note that a common requirement for a control library is to control only a subset of the degrees of freedom of a robot, e.g. control only the lower body of a legged robot while walking.
For this reason, the library must correctly compute the kinematics and dynamics of the whole system, while considering the possibility to expose only a subset of the quantities as requested by the control software.

% subsubsection model (end)

% section software_architecture (end)

%!TEX root = ../wbi.tex
\section{Implementation} % (fold)
\label{sec:software_implementation}

  % \label{fig:sw_architecture}

% \todo[inline]{``Component'' is a quite overloaded concept in Robotics Software Engineering. See
% \cite{brugaliComponent2009,brugaliComponent2010} . Typically it indicates the minimum "block" of computations available in a given robotics framework, for example OROCOS components or ROS Nodes. YARP does not have a real "component", in several aspects both a YARP module or a YARP device could be considered a "Component". Anyhow I don't think we can call this different parts of the libraries "Component".  }

This section describes the current implementation of the whole-body abstraction library conceptually described in Section~\ref{sec:software_architecture}.
The code has been implemented in C++ because of its diffusion and computational performance while remaining a high-level programming language.
The implementation has been divided in two libraries: the \emph{wholeBodyInterface} \cite{wbi} and \emph{yarpWholeBodyInterface} libraries \cite{yarpWBI}.

\subsection{wholeBodyInterface} % (fold)
\label{sub:wholebodyinterface}
The \emph{wholeBodyInterface} is the direct transposition in C++ of the abstract concepts described in Section~\ref{sec:software_architecture}.
The four elements, i.e. \emph{actuators}, \emph{sensors}, \emph{state} and \emph{model}, are represented as abstract classes.
Additionally, the library provides utilities to identify the various degrees of freedom.

As it represents the coded counterpart of the abstraction library, \emph{wholeBodyInterface} does not make any assumption on the underlining robot framework, or how data is organized, using only native C++ types.

% subsection wholebodyinterface (end)

\subsection{yarpWholeBodyInterface} % (fold)
\label{sub:yarpwholebodyinterface}

The \emph{yarpWholeBodyInterface} is the actual implementation of \emph{wholeBodyInterface} specifically considering YARP-powered mechanical systems \cite{metta2006yarp}.
Regarding the \emph{model} implementation we choose as kinematic and dynamic library the iDynTree library \cite{Frontiers2015} and information about the kinematic and dynamic model can be loaded from different sources, e.g. a URDF representation.

The \emph{actuators} and \emph{sensors} elements directly interact with YARP control boards.
Because a robot possesses in general multiple control boards, these two elements are also responsible for mapping the information coming from the control boards to the degrees of freedom selected by the library user.

Note that, because of the dependency on the YARP library, in the current implementation the \emph{state} element uses YARP data structures, e.g. vectors and matrices, but this dependency can be easily dropped in future implementations.

% subsection yarpwholebodyinterface (end)

\subsection{Simulink Interface for Model-Driven Engineering} % (fold)
\label{sub:simulink_interface}

C++ applications can leverage the advantages of the proposed abstraction layer while keeping full control of the performance of the control software by directly using the provided C++ implementations.
On the other hand, coding and testing a complex control system directly in C++ can be prohibitive.
For example, even the simple task of monitoring a signal over time can be complex and requires the use of a dedicated library.
The use of software to design and simulate dynamical system models greatly helps the design and synthesis of control systems. 
Domain-specific software for dynamical systems is a specific case of model-driven engineering \cite{brugali2015}.

We currently implemented the Simulink interface to our proposed whole-body abstraction library, which can be found in \cite{WBToolbox}.
Most of the features accessible in C++ are also accessible to Simulink models.
Furthermore, because the connection with the robot or the simulator is handled by the underlining C++ library, the Simulink interface does not require any particular toolbox to command the robot, e.g. Simulink Real Time\textsuperscript{\textregistered}.

A further advantage of using Simulink\textsuperscript{\textregistered} with respect to the C++ code consists in the possibility to exploit the abundance of toolboxes and Matlab native functions out of the box.

% subsection simulink_interface (end)

% section software_implementation (end)

%!TEX root = ../wbi.tex
\section{Experiments}
\label{sec:experiments}

This section presents the implementation of the PD plus gravity compensation controllers briefly described in Section \ref{sub:control_examples}. 
We also discuss the results of a more complex controller, namely a momentum-based balancing controller which has been implemented with the Simulink interface described in Section~\ref{sub:simulink_interface}.

\subsection{PD plus Gravity Compensation} % (fold)
\label{sub:pd_plus_gravity_compensation}

This section reports the code for the  example presented in Section~\ref{sub:control_examples}, i.e. the code for the PD plus gravity compensation controller.

\begin{figure*}[t]
  \centering
  	\def\svgwidth{\textwidth}
    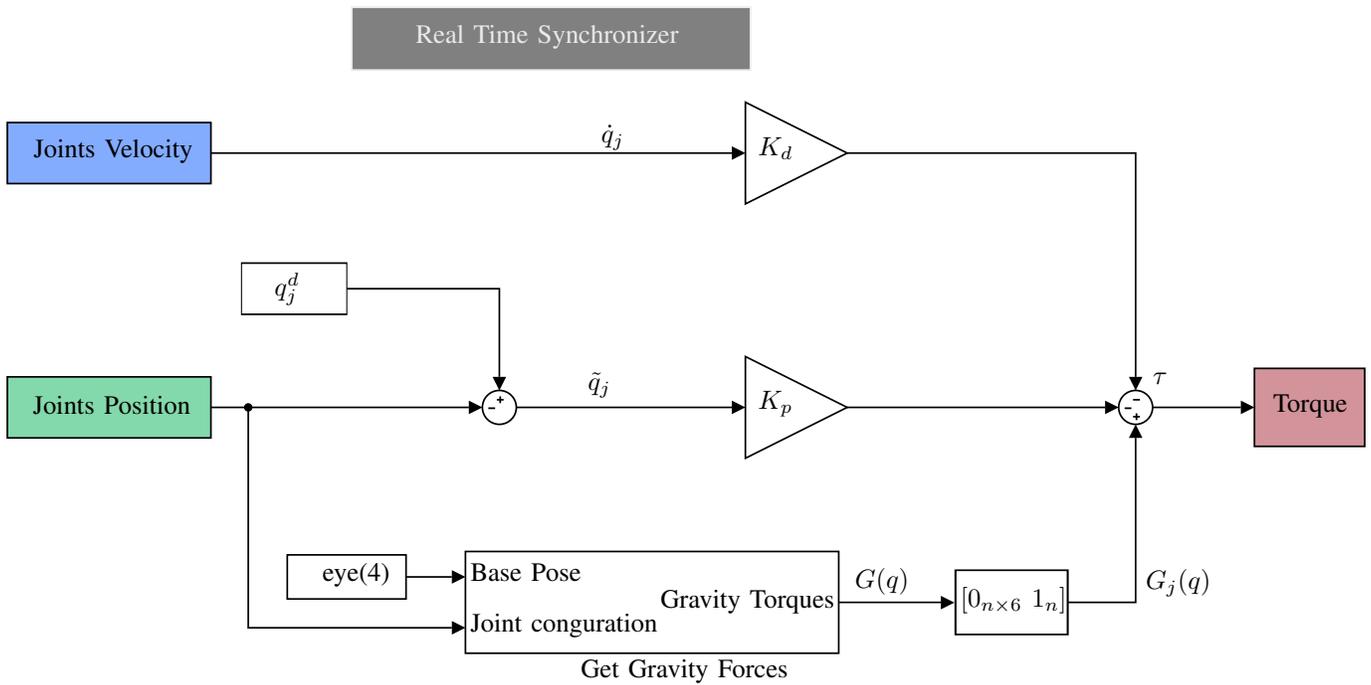
  \caption{Simulink model diagram of the PD plus gravity compensation controller for a fixed-base robot}
  \label{fig:figs_PD_plus_grav_simulink}
\end{figure*}

Because it is a simple example we show both the C++ code (see Code~\ref{code:wbi_init} and \ref{code:cpp_pd_plus_grav}) and the Simulink model diagram (see Figure~\ref{fig:figs_PD_plus_grav_simulink}).
Note that, while the Simulink diagram completely represents the controller, the C++ code snippet has been extracted from the main loop function, i.e. the function which runs at every iteration. 
How the control thread is created and managed depends on the particular system and it is outside the scope of the present paper.

The snippet of code in Code~\ref{code:wbi_init} shows how the specific YARP-based implementation is instantiated. 
In particular, the current implementation needs information about the URDF model representing the kinematic and dynamic information of the robot and the mapping between the model joints and the YARP control boards. This is provided by the object created at line $4$ and passed to the interface constructor at line $7$.
Additionally, the list of controlled joints are passed to the interface at line $19$, just before the interface initialization routine is called.

Reading the code in Code~\ref{code:cpp_pd_plus_grav} it is possible to observe how all the details regarding the specific robot platform are hidden by the library.
The object {\tt robot}, in fact, is accessed through its abstract type, as it can be also seen during its instantiation, i.e. in line $7$ of Code~\ref{code:wbi_init}.
In lines $4-7$ the state of the robot, i.e. $(q_j, \dot{q}_j)$, is read.
The feedforward term, corresponding to $G(q)$ is computed at lines $10 - 14$ where the last parameter is the resulting gravity compensation term.
Finally the error and the feedback term necessary to implement Eq.~\eqref{eq:pd_plus_grav_law} is computed in lines $17-22$. 
Because we did not use any specific mathematical library we explicitly computed the term $K_p \tilde{q}_j + K_d \dot{q}_j$ in the {\tt for} loop.
Finally, at line $25$ we send the torque command to the robot, which we previously setup to be controlled in torque mode.

\begin{algorithm}
    \centering
\begin{lstlisting}
//Properties.
// - Fill with model URDF path
// - Yarp controlboard mapping
yarp::os::Property wbiProperties = ...;

//create an instance of wbi
wbi::wholeBodyInterface* m_robot =
new yarpWbi::yarpWholeBodyInterface(
    "PD plus gravity", 
    wbiProperties);
    
if (!m_robot) {
    return false;
}

//Create list of controllable joints
wbi::IDList controlledJoints = ...;

m_robot->addJoints(controlledJoints);
if (!m_robot->init()) {
    return false;
}

\end{lstlisting}
\caption{C++ code snippet for library initialization}
\label{code:wbi_init}
\end{algorithm}

\begin{algorithm}
    \centering
\begin{lstlisting}
  wbi::Frame w_H_b; //identity + zero vector
  
  //read state
  (*@\textcolor{wbi_position}{robot->getEstimates(wbi::ESTIMATE\_JOINT\_POS,}@*)
                      (*@\textcolor{wbi_position}{positions);}@*)
  (*@\textcolor{wbi_velocity}{robot->getEstimates(wbi::ESTIMATE\_JOINT\_VEL,}@*)
                      (*@\textcolor{wbi_velocity}{velocities);}@*)
  
  //use model to compute feedforward
  robot->computeGravityBiasForces(
                      positions, 
                      w_H_b, 
                      grav,
                      gravityCompensation);
  
  //compute feedback.
  for (int i = 0; i < robot->getDoFs(); i++) {
      error(i)   = positions(i) - reference(i);
      torques(i) = gravityCompensation(i + 6) 
                 - kp(i) * error(i) 
                 - kd(i) * velocities(i);
  }
  
  //send desired torques to the robot
  (*@\textcolor{wbi_torque}{robot->setControlReference(torques);}@*)

\end{lstlisting}
\caption{C++ code for PD plus Gravity compensation}
\label{code:cpp_pd_plus_grav}
\end{algorithm}

Figure~\ref{fig:figs_PD_plus_grav_simulink} shows the same code implemented directly in Simulink.
It is evident how the block-based diagram is clearer with respect to its C++ counterpart.
Furthermore, the possibility to add scopes, or dump signal variables directly into Matlab workspace greatly increases its advantages with respect to directly coding in C++.

% subsection pd_plus_gravity_compensation (end)

\subsection{Momentum-based Balance Control} % (fold)
\label{sub:subsection_name}

To show the power of the proposed architecture we present here a second example, i.e. we show the results of a momentum-based balancing controller which has been synthesized directly by using the Simulink interface. 
Given the complexity of the control problem we do not report here screenshots or code snippets of the Simulink model, but the model can be examined in \cite{WBTController}, while the mathematical formulation can be found in \cite{nava16}.

The YouTube\textsuperscript{\textcopyright} video \cite{iCubWithSim} shows the robot performing complex movements by using the controller implemented and running as a Simulink Model.
By using the \emph{yarpWholeBodyInterface} implementation we also leverage the capabilities of the YARP middleware to seamlessly connect to the real or simulated system.
In particular the test platform is the iCub humanoid robot \cite{Metta20101125}, endowed with $53$ degrees of freedom, 6-axis force/torque sensors and distributed tactile skin.
The robot is simulated on the Gazebo simulator \cite{Koenig04} by means of Gazebo-YARP plugins \cite{YarpGazebo2014}.
The same demo has also been implemented on a different configuration of the iCub platform \cite{UtubeHeiCub}.
Note that the two robots have a different set of degrees of freedom.
Thanks to the flexibility of the library, the controller code remains the same in both scenarios.

We encourage the interested reader to test the controller on the Gazebo Simulator. 
Instructions on how to run the controller can be found directly in the model repository {\tt readme} \cite{WBTController}.
% subsection subsection_name (end)
%!TEX root = ../wbi.tex
\section{Conclusions}
\label{sec:conclusions}

In this paper we presented a software abstraction layer to simplify the development of whole-body controllers.
While there are already some whole-body control software libraries, they already define the controller structure and leave to the user only the possibility to specify objectives and constraints.

On the other hand the proposed library leaves complete freedom to the control designer by exposing all the information needed. It does not make any assumptions on the controller structure.
The whole-body abstraction library presents also the following advantages:
\begin{itemize}
    \item it decouples the writing of the controller from a particular robot implementation
    \item it decouples the writing of the controller from a specific dynamic library implementation
    \item it allows more concise and clear code as it represents uniquely the code needed to implement the mathematical formulation of the controller. All the implementation details are left to the library
    \item it allows to benchmark the controller on different platforms or with different implementations.
\end{itemize}
Furthermore, the possibility to expose the functionality at an higher level than C++ facilitates the writing of controllers as the results on the iCub robot clearly prove.

We voluntarily did not consider some aspects as they are out of the scope of the present contribution. 
Nevertheless they must be taken into account when a controller is implemented and used on the real system.
In particular the following details should be considered:
\begin{itemize}
    \item how are controllers run on the platform? Do they run as threads?
    \item how are controllers configured and initialized?
    \item how is communication with other software performed? For example, how are desired values provided to the controller, coming from a planner or higher-level control loop?
\end{itemize}
By not considering these details in the abstraction library, we render the library portable to different systems.
Indeed, the actual control law is not concerned by the previously listed implementation details.

While the more complex demos have been achieved by directly executing the Simulink model connected to the robot, we recognize the need to automatically generate self-contained C++ code.
The advantage is twofold.
On one side the autogenerated code is in general more optimized than the code directly executed in Simulink, even if less optimized than ad-hoc C++ code.
On the other side, this would remove the requirement of having a Simulink installation on the computers controlling the robot.



\bibliographystyle{IEEEtran}
\bibliography{IEEEabrv,Bibliography}

\end{document}